\documentclass[10pt,twocolumn,letterpaper]{article}

\usepackage{iccv}
\usepackage{times}
\usepackage{epsfig}
\usepackage{graphicx}
\usepackage{amsmath}
\usepackage{amssymb}
\usepackage{multirow}
\usepackage{caption}
\usepackage{subcaption}
\usepackage{pgfplots}
\usepackage{colortbl}
\usepackage{bbm}
\usepackage{enumitem}
\usepackage{bbding}
\usepackage{booktabs}
\definecolor{mygray}{gray}{0.9}
\usepackage[pagebackref=true,breaklinks=true,letterpaper=true,colorlinks,bookmarks=false]{hyperref}

\iccvfinalcopy 

\def\iccvPaperID{****} 

\ificcvfinal\fi

\begin{document}

\title{Random Boxes Are Open-world Object Detectors}

\author{%
 \textbf{Yanghao Wang}\textsuperscript{1}, \quad \textbf{Zhongqi Yue}\textsuperscript{1,2}, \quad \textbf{Xian-Sheng Hua}\textsuperscript{3}, \quad \textbf{Hanwang Zhang}\textsuperscript{1}\\
\small \textsuperscript{1}Nanyang Technological University,\quad \textsuperscript{2}Damo Academy, Alibaba Group,\quad \textsuperscript{3}Terminus Group\\
\tt\small scuwyh2000@gmail.com,
 yuez0003@ntu.edu.sg, \tt\small huaxiansheng@gmail.com, hanwangzhang@ntu.edu.sg\\}

\maketitle
\ificcvfinal\thispagestyle{empty}\fi

\begin{abstract}
We show that classifiers trained with random region proposals achieve state-of-the-art Open-world Object Detection (OWOD): they can not only maintain the accuracy of the known objects (w/ training labels), but also considerably improve the recall of unknown ones (w/o training labels). Specifically, we propose RandBox, a Fast R-CNN based architecture trained on \underline{random proposals} \underline{at each training iteration}, surpassing existing Faster R-CNN and Transformer based OWOD. Its effectiveness stems from the following two benefits introduced by randomness. First, as the randomization is independent of the distribution of the limited known objects, the random proposals become the instrumental variable that prevents the training from being confounded by the known objects. Second, the unbiased training encourages more proposal explorations by using our proposed matching score that does not penalize the random proposals whose prediction scores do not match the known objects. On two benchmarks: Pascal-VOC/MS-COCO and LVIS, RandBox significantly outperforms the previous state-of-the-art in all metrics. We also detail the ablations on randomization and loss designs. Codes are available at \href{https://github.com/scuwyh2000/RandBox}{https://github.com/scuwyh2000/RandBox}.
\end{abstract}
\section{Introduction}
\label{sec:1}

\begin{figure}[tbp]
	\centering
	\subfloat
 {\includegraphics[width=1\linewidth]{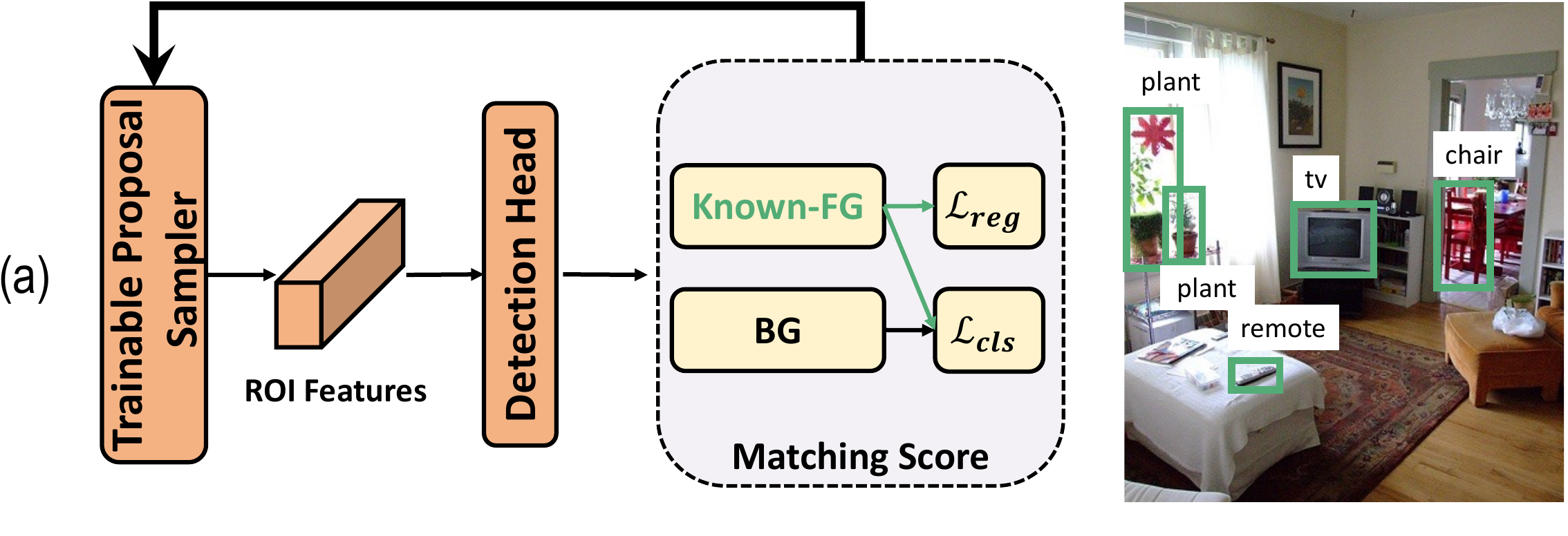}}\\

	\subfloat{\includegraphics[width=1\linewidth]{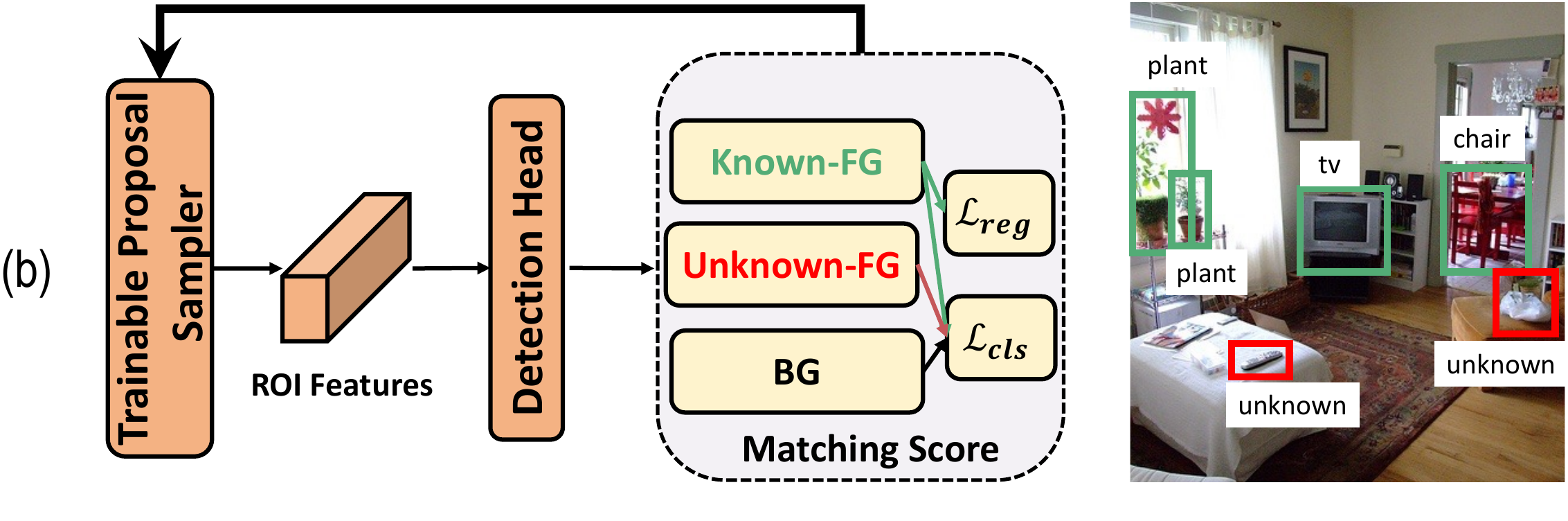}}\\	

        \subfloat{\includegraphics[width=1\linewidth]{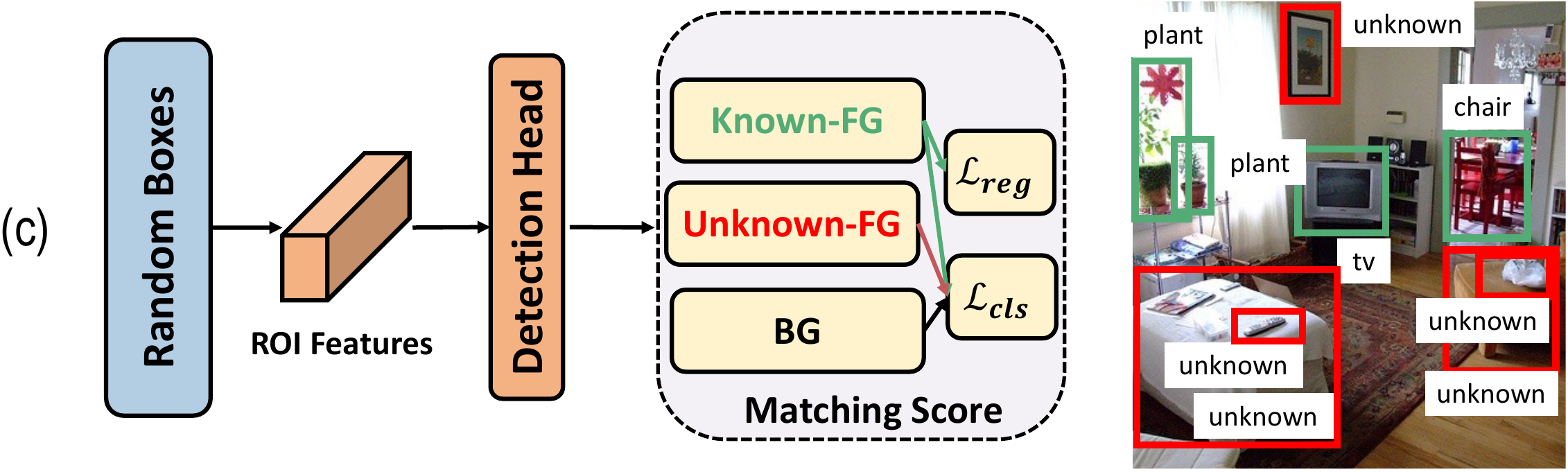}}\\
	\caption{The paradigm of (a) closed-world object detection, (b) existing open-world object detection, and (c) our RandBox that uses random proposals.}
 \label{fig:1}
\end{figure}

Different objects may share common visual features. For example, the detection of ``horse'' helps that of ``cow'' as they are both ``four-legged'' cattle. Therefore, it is possible to  generalize the detector trained from a limited object class inventory to detect visually similar objects whose classes are out of the inventory.  As shown in the right column of Figure~\ref{fig:1}, besides the conventional detection of a close-world object inventory, Open-wolrd Object Detection~\cite{joseph2021towards} (OWOD) can detect out-of-inventory objects and label them as ''unknown''.

To exploit the feature transfer from ``known'' to ``unknown'', the training loss of existing OWOD methods~\cite{joseph2021towards, gupta2022ow, yang2021objects} has the following break-downs (Figure~\ref{fig:1}):

\noindent\textbf{Known-FG (foreground)}: like conventional closed-world detection (Figure~\ref{fig:1}a), we calculate the matching scores between the predicted region proposal labels and the ground-truth labels of known objects, where the label is a pair of class and bbox (bounding box). In particular, for two-stage detectors like Faster R-CNN~\cite{ren2015faster}, the ``proposal sampler'' and ``matching score'' are region proposal network and bbox IoU (intersection over union); for end-to-end detectors like DETR~\cite{carion2020end}, they are query transformer decoder and bipartite matching score. For the matched proposals, the loss is class cross-entropy $\mathcal{L}_{cls}$ and  bbox regression $\mathcal{L}_{reg}$.

\noindent\textbf{BG (background)}: for conventional detection, all the mismatched proposals are considered background, whose loss is only $\mathcal{L}_{cls}$ with the label ''BG''; for OWOD, we only use the mismatched proposals excluding the Unknown-FG proposals below.

\noindent\textbf{Unknown-FG}: this is the key difference from conventional detection. Its design heuristic is that, for the mismatched proposals with class confidence larger than a  threshold---probably unknown objects---the loss is only $\mathcal{L}_{cls}$ with a special label ``unknown'' w/o bbox regression. Note that this heuristic is based on the assumption of feature transfer~\cite{zheng2016good,chen2018lstd}---visually similar proposals have similar class confidences (see Section~\ref{sec:3.1} for the review of the loss implementation in two-stage and end-to-end detection)---the more unknown objects the Unknown-FG collects, the better the OWOD.
\begin{figure}[t!]
\resizebox{\linewidth}{!}{
\centering

\includegraphics[]{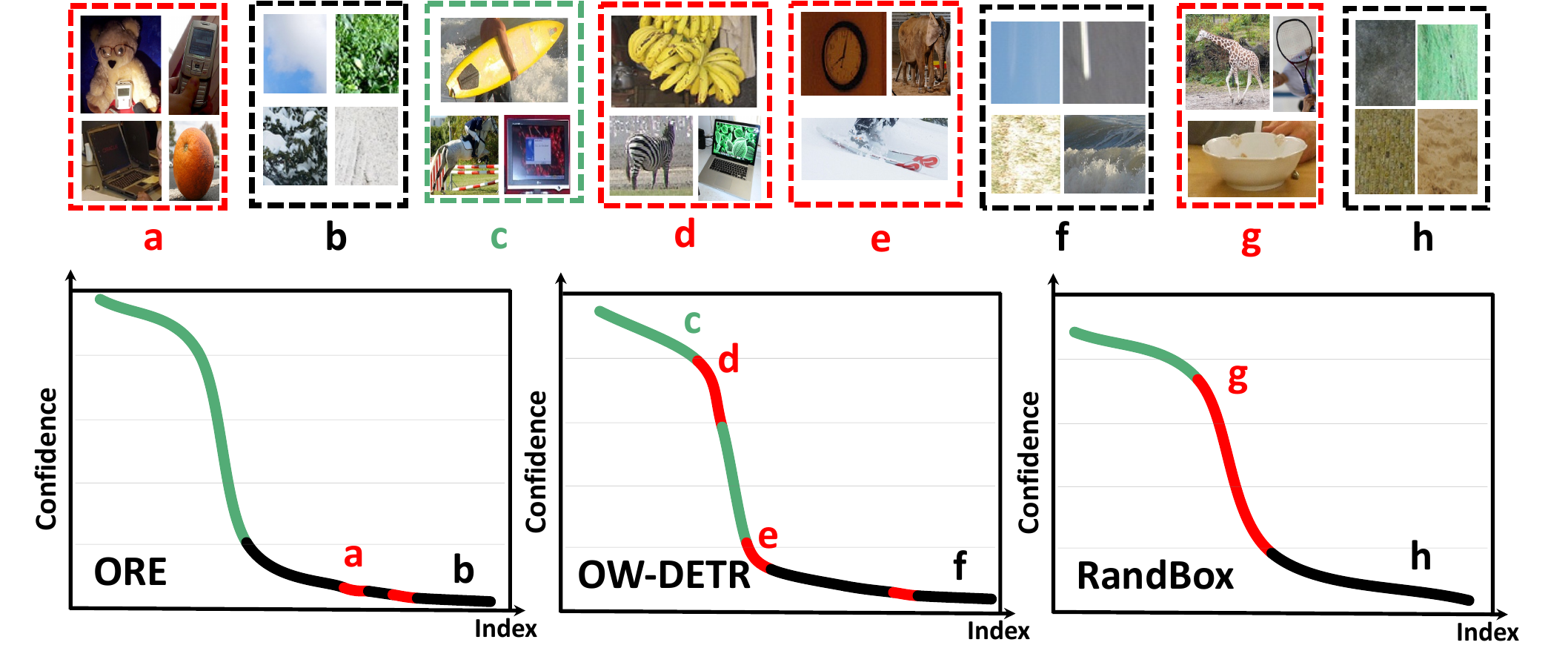}
\centering
}
\caption{The sorted foreground confidence assigned by two baselines (ORE~\cite{joseph2021towards}, OW-DETR~\cite{gupta2022ow}) and our RandBox to the generated proposals in training. Green, red and black correspond to ``known'', ``unknown'' and ``BG'', respectively, with examples showing on the top row.}
\label{fig:3}
\end{figure}

However, in practice, Unknown-FG usually has a low recall of unknown objects. As shown in Figure~\ref{fig:3}, in existing methods, most proposals of high confidence are matched as Known-FG; the mismatched proposals, mixed with both unknown objects and BG, all have low confidence, losing a gradual confidence decrease to differentiate Unknown-FG from BG (the red and black colors are inseparable). One may easily understand that this failure is predictable because, yet, we don't have the ground-truth labels for unknown objects, so the proposal sampler trained on known objects will be inevitably biased towards them. In fact, such bias is ubiquitously in any bootstrapping learning~\cite{riloff2003learning} with limited supervision, such as semi-supervised learning~\cite{miyato2018virtual,van2020survey} and reinforcement learning~\cite{kaelbling1996reinforcement,wiering2012reinforcement}. Unfortunately, existing OWOD methods do not properly address the bias. 

\begin{figure*}[htbp]
\resizebox{\linewidth}{50mm}{
\centering

\includegraphics[]{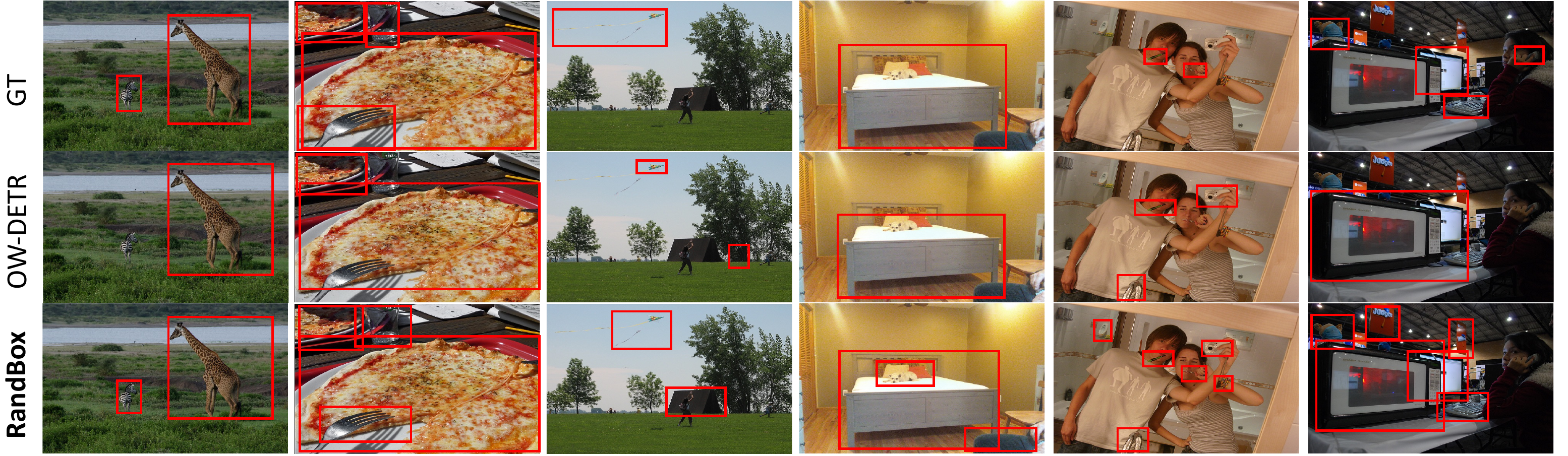}
\centering
}
\caption{\textbf{Example results}. We show the ground-truth (GT) locations of unknown objects, those predicted by OW-DETR~\cite{gupta2022ow}, and those predicted by our RandBox. RandBox not only detects the annotated unknown objects (first 2 images), but even discovers valid unknown ones without annotations (\eg, \textit{camp} in image 3, \textit{bracelet} in image 5, \textit{banner} in image 6).}
\label{fig:2}
\end{figure*}

In this paper, we point out that the bias is caused by the confounding effect~\cite{pearl2012calculus,pourhoseingholi2012control} of the training data with limited known object labels that mislead the feature transfer. The effect can be understood in this way: if we only let the training data determine what region proposals to be used to extract the region features for detector training (Figure~\ref{fig:1}b), we cannot avoid the trivial loss minimization from only detecting the known objects. To this end, we propose to remove the confounding effect by introducing a random Instrumental Variable~\cite{baiocchi2014instrumental}---the proposed RandBox---that is independent of the training data. Intuitively, RandBox simulates a Randomized Controlled Experiment (RCT)~\cite{braga1999problem} to learn a causal detector from region features to object labels,  by inviting an independent 3rd-party proposal sampler to explore more possible locations of unknown objects. As shown in Figure~\ref{fig:3} RandBox, the recall of unknown objects is greatly improved by RandBox, due to the fact that the confidence score change of the proposals from known to unknown is much more separable. We will detail the theory behind RandBox in Section~\ref{sec:4}. To get a qualitative sense of open-world detection task, see Figures \ref{fig:2}.

RandBox is simple yet effective (Section~\ref{sec:3.2}). For each training image in every iteration, we sample 500 random boxes as proposals from a 4-dimensional Gaussian noise (center\_x, center\_y, height, width). For these random boxes, we crop their RoI (region of interest) features from a backbone feature map, and feed them into a detection head to obtain $\mathcal{L}_{cls}$ and $\mathcal{L}_{reg}$ as in (Figure~\ref{fig:1}c), where the proposed matching score that encourages exploration is detailed in Section~\ref{sec:3.2}. At inference, we remove the randomness and grid-sample 10,000 boxes per image by 10 scales, 10 shapes, and 100 spatial locations. We use NMS~\cite{neubeck2006efficient} (Non-maximum Suppression) to prune the 10,000 boxes for final detection. Overall, our contributions include:
\begin{itemize}[leftmargin=+0.25in,itemsep=3pt,topsep=3pt,parsep=0pt]
    \item Our proposed RandBox is a Fast R-CNN based architecture for OWOD, which leverages random proposals to remove the confounding effect from the known objects, and uses an improved matching score to encourage proposal exploration.
    \item By removing the confounding effect, we learn an unbiased detector that improves both known and unknown detection accuracy. In particular, we achieve new state-of-the-art results on Pascal-VOC/MS-COCO~\cite{lin2014microsoft}.
    \item We show that our RandBox detector trained on Pascal-VOC/MS-COCO remains robust when transferring to long-tailed datasets such as LVIS~\cite{gupta2019lvis}.
\end{itemize}
\section{Related Work}
\label{sec:2}
\noindent\textbf{Open-World Object Detection (OWOD)}.
Object detection systems are significantly advanced recently~\cite{redmon2016you,zhu2020deformable,ren2015faster,girshick2015fast,sun2021sparse}. Yet they are based on the conventional close-world paradigm, which assumes that all objects of interests are annotated. In practice, an object detector deployed in real life is constantly challenged by the vast open and dynamic visual world~\cite{miller2018dropout,joseph2021towards}.
To bridge the gap, open-set detection~\cite{miller2018dropout,miller2019evaluating,dhamija2020overlooked} aims to reject out-of-inventory objects as ``unknown'', and other works focus on incrementally learning~\cite{rajasegaran2020itaml,aljundi2018memory,mallya2018packnet} a detector to recognize new object classes without forgetting old ones.
%
Inspired by the open-world setting proposed in image classification~\cite{bendale2015towards,bendale2016towards}, recent works~\cite{joseph2021towards,gupta2022ow,yang2021objects} unify open-set detection and incremental learning as OWOD, which are based on two stage detectors like Faster R-CNN~\cite{ren2015faster}, or end-to-end ones like DETR~\cite{zhu2020deformable}.
We systematically analyze the failure of these existing works---low recall of unknown objects---which is caused by the confounding effect~\cite{pearl2012calculus,pearl2009causality}, and remove it using random proposals to learn a causal detector.

\noindent\textbf{Debiasing by Randomization}.
Randomization is an effective tool to debias across many tasks and domains.
For example, using randomization as an instrument variable~\cite{deaton2010instruments,arellano1995another,baiocchi2014instrumental} is a classic tool to identify causal effect in social studies~\cite{didelez2007mendelian}.
In reinforcement learning~\cite{wiering2012reinforcement}, randomization is used to encourage exploration~\cite{wilson2021balancing,kveton2020randomized}, such that the agent is not biased to exploiting spurious shortcuts.
Several works~\cite{araujo2019robust,wang2021augmax} in adversarial training~\cite{ganin2016domain} leverages randomization to build a robust model not biased to feature that can be manipulated in adversarial attack, \eg, by image pertubation~\cite{zwanenburg2019assessing}.
In fact, data augmentation that is prevailing in vision~\cite{shorten2019survey,he2022masked,sohn2020fixmatch} and language~\cite{shorten2021text,salazar2019masked} also leverages randomness to remove augmentation-related context bias in training (\eg, using green grass to classify cow).
In our work, we present a novel detection framework that leverages random proposals to learn causal effect in OWOD.

\section{Method}
\label{sec:3}

\begin{figure*}[htbp]

\resizebox{\linewidth}{!}{
\centering

\includegraphics[]{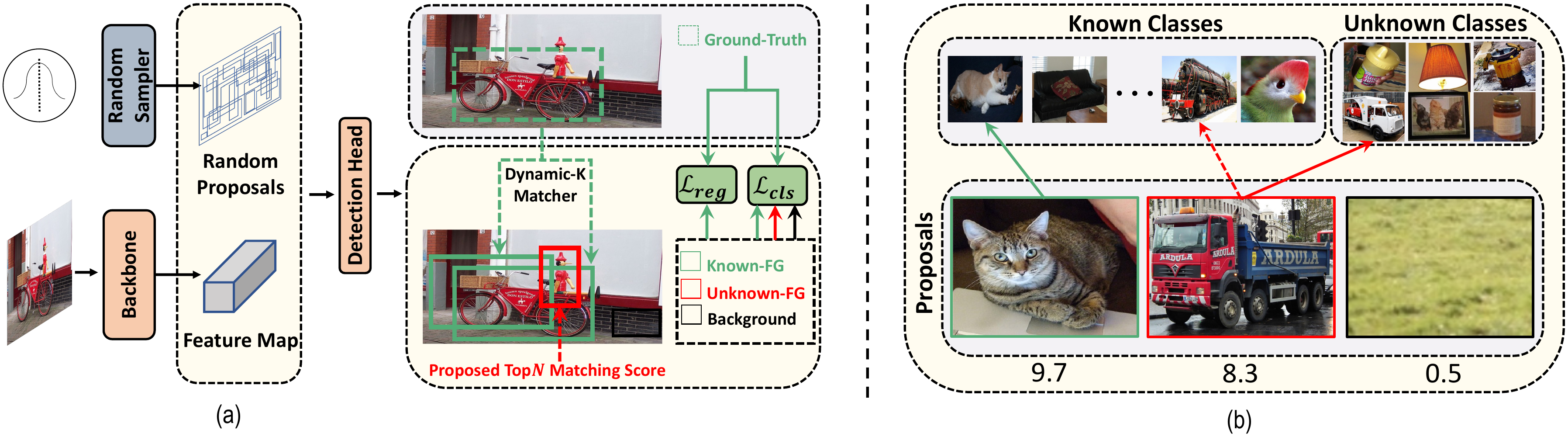}
\centering
}

\footnotesize

\caption{(a) \textbf{RandBox Pipeline.} Our proposals are randomly generated. We obtain the Known-FG subset using a dynamic-$k$ matcher, and the Unknown-FG subset based on our proposed matching score. Our training losses follow existing methods. (b) \textbf{Proposed Matching Score} is the sum of likelihood of belonging to each foreground classes (including ``unknown''). The proposals of known (green) and unknown (red) objects both have large matching score, \ie, high likelihood on the ground-truth class (solid arrows) or by feature transfer (dashed arrow), while BG proposals (black) have low score.}
\label{fig:4}
\end{figure*}

In OWOD, the task is to detect a gradually expanding set of known inventory classes $K$, while simultaneously rejecting the out-of-inventory classes set $U$ as ``unknown''.
The training dataset contains images where known objects are labelled.
Specifically, for a training image containing $n$ known objects, it is labelled with a set $\{(b_i,y_i)\}_{i=1}^n$, where $b_i$ specifies the bounding box (bbox) of the $i$-th object (\ie, center\_x, center\_y, height, width), and $y_i\in {K}$ is its class label.
We drop the subscript $i$ for simplicity when the context is clear.
The goal in OWOD is to train a detector, whose input is an image, and output is a set of $m$ predictions $P=\{(\hat{b}_i, \hat{y}_i)\}_{i=1}^m$, where $\hat{b}_i$ is the $i$-th bounding box prediction (predicted bbox), and $\hat{y}_i\in \mathbb{R}^{|{K}| + 2}$ denotes the classification logit for each class in ${K} \cup \{\textrm{``unknown''}, \textrm{``BG''}\}$.

\subsection{Preliminaries: Existing OWOD Methods}
\label{sec:3.1}
Existing methods adopt either the two-stage Faster R-CNN~\cite{ren2015faster} or the end-to-end DETR~\cite{carion2020end} as the detector. In training, the predictions ${P}$ generated by the detector are partitioned into three subsets: Known-FG, Unknown-FG and BG, which are used to calculate the classification loss and bbox regression loss. We detail each part below:

\noindent\textbf{Detector}. Given an image, a detector generates $m$ proposals, extracts ROI features ${x}_i$ for each proposal, and finally computes bounding box $\hat{b}_i$ and logits $\hat{y}_i$ using a classifier to form ${P}$.
The two detector frameworks mainly differ in the first step:
1) Faster R-CNN leverages a region proposal network (RPN) pre-trained on the labelled data. It outputs the top-$m$ proposals with the largest \emph{objectness} (\ie, probability of containing a known object). 2) DETR is an end-to-end detector. It first transforms an image into tokens using a combination of CNN backbone and transformer encoder. Then a transformer decoder uses the image tokens to generate the proposals. More details are in Appendix.

\noindent\textbf{Known-FG} is selected from all predictions $P$ by matching with the labels $\{(b_i,y_i)\}_{i=1}^n$ based on a Known-FG score.
To compute the score for a prediction $(\hat{b}, \hat{y}) \in P$ with a ground-truth $(b,y)$, Faster R-CNN uses the IoU between $\hat{b}$ and $b$, and DETR uses the bipartite matching score, which additionally considers if $\hat{y}$ has a large value on the class $y$.
As the process is the same as that in closed-set detection, we leave other details in Appendix.
After matching, we denote the Known-FG subset as ${P}^K$. For $(\hat{b}, \hat{y}) \in {P}^K$, we denote its matched ground-truth as $(b^*, y^*)$.

\noindent\textbf{Unknown-FG \& BG} are selected from the unmatched predictions ${P} \setminus {P}^K$. Specifically, current methods compute an Unknown-FG score for each prediction, and select $5$ predictions with the largest score as the Unknown-FG. For the score, methods based on Faster R-CNN~\cite{ren2015faster,joseph2021towards} use objectness, and those based on DETR~\cite{gupta2022ow} use the average value of the ROI feature. We denote the Unknown-FG subset as ${P}^U$. Finally, the rest predictions become the BG subset ${P}^B$.

\noindent\textbf{Training Loss}.

We calculate the the loss on each of the three subsets. The loss on Known-FG $\mathcal{L}^K$ has the same form with closed-set detection:
\begin{equation}
    \mathcal{L}^K = \sum_{i=1}^{|{P}^K|} \mathcal{L}_{cls}(\hat{y}_i, y_i^*) + \lambda \mathcal{L}_{reg}(\hat{b}_i, b_i^*),
    \label{eq:1}
\end{equation}
where $\mathcal{L}_{cls}$ is the cross-entropy loss or focal loss~\cite{lin2017focal} to handle class-imbalance, $\lambda$ is a balancing weight, and $\mathcal{L}_{reg}$ is the smooth L1 regression loss~\cite{girshick2015fast} to rectify the predicted bbox. Details of $\mathcal{L}_{cls}$ and $\mathcal{L}_{reg}$ are in Appendix.
As Unknown-FG and BG have no annotated bbox, we only calculate  $\mathcal{L}_{cls}$ on them:

\begin{equation}
    \mathcal{L}^{UB} = \sum_{i=1}^{|{P}^U|} \mathcal{L}_{cls}(\hat{y}_i, \textrm{``unknown''}) + \sum_{i=1}^{|{P}^B|} \mathcal{L}_{cls}(\hat{y}_i, \textrm{``BG''}),
\label{eq:2}
\end{equation}
Overall, the detector parameters $\theta$ are optimized with:
\begin{equation}
    \mathop{\mathrm{min}}_{\theta} \mathcal{L}^K + \beta \mathcal{L}^{UB},
    \label{eq:3}
\end{equation}
where $\beta$ is a weight with ablation in Section~\ref{sec:5.2}.

\noindent\textbf{OWOD Pipeline}.
The detector is first randomly initialized. Then it is trained with the labelled data on ${K}$, and evaluated with data on both ${K}$ and ${U}$ (Task 1). Next, additional labelled data on a subset of classes ${K}' \subset {U}$ is provided, \ie, the known classes set expands as ${K}\leftarrow {K} \cup {K}'$, and the unknown one becomes ${U}\leftarrow {U} \setminus {K}'$. The detector trained on Task 1 is fine-tuned and evaluated with the updated ${K}$ and ${U}$ (Task 2). This continues for the following tasks.

\subsection{Proposed RandBox}
\label{sec:3.2}

Existing methods suffer from the low recall of the unknown objects in Unknown-FG. To tackle this, our proposed RandBox has two improvements: 1) We propose to generate region proposals randomly, instead of using a proposal sampler trained on the known objects, which explores more possible unknown object locations. 2) We propose a matching score without penalizing proposals that are not matched with the known objects, which explores more possible unknown object proposals. We detail the two improvements below, and explain why our RandBox solves the existing problems from a causal view in Section~\ref{sec:4}.

\noindent\textbf{Detector with Random Proposals}.
In training, we randomly generate 500 bboxes as region proposals on each image (Figure~\ref{fig:4}a).
For each bbox, we sample 4 random real numbers in $[0,1]$, where each number is first drawn from the standard Gaussian distribution, truncated to $[-2,2]$ and then scaled linearly to $[0,1]$.
The 4 numbers correspond to the horizontal and vertical coordinate of the bbox center, as well as the width and height of the bbox, respectively (proportional to image size). The width and height of out-of-boundary bboxes are adjusted to stay in the image.
In testing, to remove the prediction randomness, we use 10,000 pre-defined bboxes as proposals on each image, which cover 10 scales, 10 aspect ratios and 100 spatial locations (details in Appendix). We use non-maximum suppression to prune the bboxes as per standard practice~\cite{neubeck2006efficient}.
We highlight the following points:
\begin{itemize}[leftmargin=+0.1in,itemsep=1pt,topsep=1pt,parsep=0pt]
    \item Our region proposals are randomly sampled for every image in each iteration throughout the training. 
    \item While we use random sampling, the number of our random proposals is in fact \emph{comparable} with that used in the existing methods, \eg, ORE~\cite{joseph2021towards} uses 1024 proposals.
\end{itemize}

\noindent\textbf{Known-FG}. We adopted the dynamic-$k$ matcher~\cite{chen2022diffusiondet} as shown in Figure~\ref{fig:4}a. Specifically, each ground-truth bbox $b$ is matched with $k$ proposals based on the bipartite matching score, where $k$ is dynamically selected as the sum of the IoU between each proposal and $b$.

\noindent\textbf{Unknown-FG}. We propose a novel \textbf{matching score} to form Unknown-FG. Specifically, we select the top $N$ proposals with the largest matching score (ablation of $N$ in Section \ref{sec:5.2}). Among the selected proposals, those already in Known-FG are removed to form our Unknown-FG subset.
We use the following equation to compute our matching score for a proposal with the classification logits $\hat{y}$:
\begin{equation}
    s(\hat{y}) = \sum_{i=1}^{|K| + 1} \frac{1}{1+\mathrm{exp}(-\hat{y}_i)},
    \label{eq:4}
\end{equation}
where we compute and sum the sigmoid of the $|K|$ known classes and ``+1'' unknown class, and $\hat{y}_i$ denotes the logit for the class $i$.
Note that we implement $\mathcal{L}_{cls}$ with the BCE loss~\cite{lin2017focal}.
Hence our score essentially computes how likely $\hat{y}$ corresponds to the foreground.
Specifically, the top-$N$ (largest matching score) proposals that are not in Known-FG are pseudo-labeled as ``unknown'' to train the Unknown-FG logit (Eq.~\ref{eq:2} first loss). This design is based on the assumption of feature transfer~\cite{perkins1992transfer,weiss2016survey}---similarity between Unknown-FG and Known-FG is larger than that between BG and Known-FG.
As shown in Figure~\ref{fig:4}b, an unknown object proposal can have non-trivial likelihood of belonging to a known class due to shared feature (dashed arrow), while BG proposal almost shares no feature with known. Hence future unknown object proposals can have large likelihood of belonging to ``unknown'' (solid red arrow), increasing the unknown recall. 
We have the following comparisons with existing methods:
\begin{itemize}[leftmargin=+0.1in,itemsep=1pt,topsep=1pt,parsep=0pt]
    \item Methods based on Faster R-CNN use the objectness predicted by RPN as the matching score to select Unknown-FG. However, RPN is trained to produce large scores \emph{only} on the proposals matched with the known objects, thus penalizing those unmatched ones.
    \item Without RPN, methods based on DETR use the mean activation of the ROI feature instead. However, this heuristic is extremely unreliable (Figure 3b), and unknown object proposals are often mistakenly placed in the BG subset and penalized as ``background''.
    \item In contrast, our matching score fairly evaluates each proposal by checking both known classes (first $|K|$ logits), as well as ``unknown'' ($|K|+1$-th logit). Hence unknown object proposals can be reliably selected in the Unknown-FG subset with high likelihood of belonging to ``unknown''.

\end{itemize}

\noindent\textbf{Other Details}.
Our training objective is the same as Eq.~\eqref{eq:3}.
We used ResNet-50~\cite{he2016deep} as backbone. Our RandBox is Fast R-CNN~\cite{girshick2015fast} based architecture and detection head is borrowed from Sparse R-CNN~\cite{sun2021sparse}.

\section{Theory}
\label{sec:4}
\begin{figure}
\centering
	\subfloat{\label{fig:4a}\includegraphics[width = 0.48\textwidth]{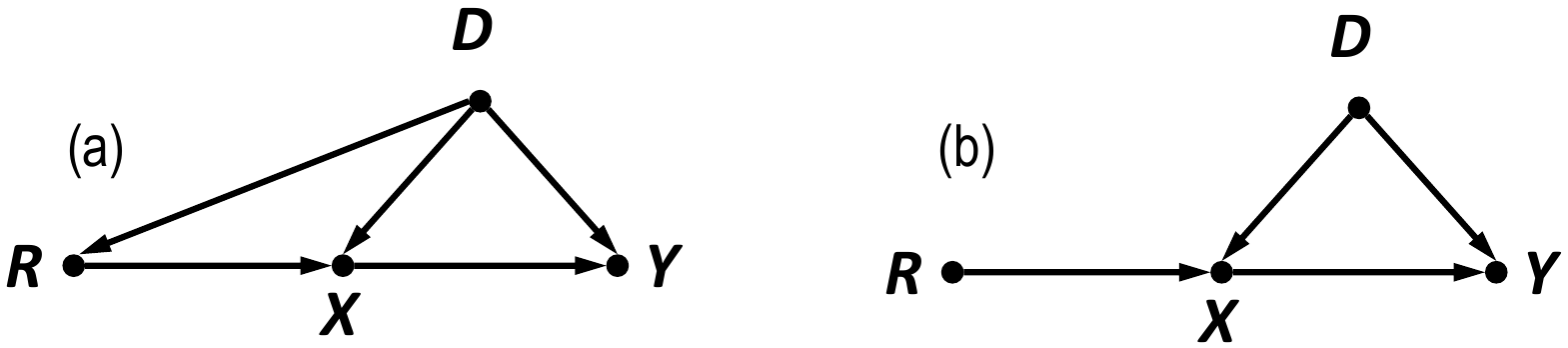}}
\caption{\textbf{Causal graph} of (a) existing OWOD methods, and (b) our RandBox.}
\label{fig:5}
\end{figure}

We use a Structural Causal Model (SCM) to analyze the causalities between the key components in OWOD: training data $D$, region proposal $R$, ROI feature $X$, and label $Y$ on ${K} \cup \{\textrm{``unknown''}, \textrm{``BG''}\}$. Note that $D$ only contains labelled data of the known classes $K$. The SCM is illustrated with a causal graph in Figure~\ref{fig:5}, where each directed link denotes a causal relation between the two connected nodes.

\noindent$\boldsymbol{R}\to \boldsymbol{X} \to \boldsymbol{Y}$ represents the desired causal effect from region proposal $R$ to label $Y$, as label $Y$ describes what a region $R$ (or its ROI feature $X$) contains. A detector is unbiased if it identifies $R\to X\to Y$, \ie, given any region, the detector can fairly evaluate what it contains.

\noindent$\boldsymbol{R} \to \boldsymbol{X} \leftarrow \boldsymbol{D}$. The ROI feature $X$ is jointly determined by what the region contains ($R \to X$), as well as the feature representation learned on the training data ($D\to X$).

\noindent$\boldsymbol{D}\to \boldsymbol{Y}$ is because the detector leverages the training knowledge obtained from $D$ to predict $Y$, \eg, identifying the BG subset with prediction ``BG'' (Section~\ref{sec:3.1}).

\noindent$\boldsymbol{D}\to \boldsymbol{R}$ exists in current OWOD methods as shown in Figure~\ref{fig:5}a. This is because the region proposal network in Faster R-CNN and the transformer decoder in DETR are both trained on $D$ to generate proposal $R$.

\noindent\textbf{Confounding Effect}. In Figure~\ref{fig:5}a, $R\leftarrow D \to Y$ is known as a backdoor path, which leads to the undesired confounding effect~\cite{pearl2012calculus}. Specifically, as the proposal sampler is supervised by the labelled data of known objects $D$, the generated proposals $R$ are inevitably biased to them ($D\to R$), \ie, $R$ only faithfully reflects the known object locations.
Consequently, the training is dominated by the loss $\mathcal{L}^K$ on Known-FG, and the detector recklessly exploits \emph{any} discriminative feature in $D$ towards the known objects ($D\to Y$), which may not transfer to the unknown ones.
This causes the low recall of Unknown-FG in Figure~\ref{fig:3}.

\begin{table*}[t!]
\centering
\scalebox{0.72}{
\def\arraystretch{1.1}
\begin{tabular}{p{3.7cm}p{1.1cm}<{\centering}p{0.7cm}<{\centering}p{0.8cm}<{\centering}p{1.1cm}<{\centering}p{0.1cm}<{\centering}p{1.1cm}<{\centering}p{0.7cm}<{\centering}p{0.8cm}<{\centering}p{1.1cm}<{\centering}p{0.1cm}
<{\centering} p{1.1cm}<{\centering}p{0.7cm}<{\centering}p{0.8cm}<{\centering}p{1.1cm}<{\centering}p{0.1cm}<{\centering} p{1.1cm}<{\centering}}
\hline\hline
\multicolumn{1}{c}{\multirow{2}{*}{\large{Method}}}    & \multicolumn{4}{c}{\textbf{Task 1}} &       &      \multicolumn{4}{c}{\textbf{Task 2}} & & \multicolumn{4}{c}{\textbf{Task 3}} & & \multicolumn{1}{c}{\textbf{Task 4}} \\ \cmidrule(lr){2-5}\cmidrule(lr){7-10}\cmidrule(lr){12-15}\cmidrule(lr){17-17}  
 & K-mAP   & U-R    & WI & A-OSE  &    & K-mAP   & U-R    & WI & A-OSE  &    & K-mAP   & U-R    & WI & A-OSE  &  & K-mAP  \\ \hline
 Faster-RCNN~\cite{ren2015faster} & 
  56.4 &
  - &
  0.0699 &
  13,396 &
  &15.2 & 
  - &
  0.0371 &
  12,291 &
  &6.7 & 
    - &
  0.0213 &
  9,174 &
  &4.2 \\
 Faster-RCNN+Finetuning& 56.4 &
  - &
  0.0699 &
  13,396 &
  &38.0 &
    - &
  0.0375 &
  12,497 &&
  30.0 &
    - &
  0.0279 &
  9,622 &&
  25.6 \\
 DDETR~\cite{zhu2020deformable} & 
  60.3 &
    - &
  0.0608 &
  33,270 &
  &17.9 &
  - &
  0.0368 &
  18,115 &
  &8.5 &
    - &
  0.0197 &
  9,392 &
  &6.0 \\
 DDETR+Finetuning& 60.3 &
  - &
  0.0608 &
  33,270 &
  &44.8 &
    - &
  0.0337 &
  17,834 &
  &33.3 &
    - &
  0.0195 &
  10,095 &
  &29.4 \\
  \hline
 ORE~\cite{joseph2021towards} & 
  56.0 &
    4.9 &
  0.0621 &
  10,459 &
  &39.4 &
    2.9 &
  0.0282 &
  10,445 &
  &29.7 &
    3.9 &
  0.0156 &
  6,803 &
  &25.3 \\
     OST~\cite{yang2021objects} & 
  56.2 &
    - &
  0.0417 &
  4,889 &
  &39.9 &
    - &
  0.0213 &
  2,546 &
  &29.6 &
    - &
  0.0146 &
  2,120 &
  &25.9 \\
      OW-DETR~\cite{gupta2022ow} &
  59.3 &
    7.5 &
  0.0571 &
  10,240 &
  &42.9 &
    6.2 &
  0.0278 &
  8,441 &
  &30.8 &
    5.7 &
  0.0156 &
  6,803 &
  &27.8 \\
      \textbf{RandBox} 
     &
  \cellcolor{mygray}\textbf{61.8} &
    \cellcolor{mygray}\textbf{10.6} &
  \cellcolor{mygray}\textbf{0.0240} &
  \cellcolor{mygray}\textbf{4,498} &
  \cellcolor{mygray}&
  \cellcolor{mygray}\textbf{45.3} &
  \cellcolor{mygray}\textbf{6.3} &
  \cellcolor{mygray}\textbf{0.0078} &
  \cellcolor{mygray}\textbf{1,880} &
  \cellcolor{mygray} &
  \cellcolor{mygray}\textbf{39.4} &
    \cellcolor{mygray}\textbf{7.8} &
  \cellcolor{mygray}\textbf{0.0054} &
  \cellcolor{mygray}\textbf{1,452} &
  \cellcolor{mygray}
  &\cellcolor{mygray}\textbf{35.4} \\
  \hline
\end{tabular}}
\caption{\textbf{OWOD Evaluation} results and comparison with existing methods.}
\label{tab:1}
\vspace{-0.3cm}
\end{table*}
\noindent\textbf{Removing Confounding Effect}. In our RandBox, $R$ is randomly generated with no influence from $D$, \ie, $D\not\to R$ in Figure~\ref{fig:5}b.
This makes $R$ an Instrument Variable~\cite{baiocchi2014instrumental}, which removes the confounding effect by simulating a Randomized Controlled Experiment~\cite{braga1999problem}.
We provide a three-fold intuitive explanation:
1) The backdoor path is cut, \ie, $R\not\leftarrow D \to Y$;
2) The path $R \to X \leftarrow D \to Y$ is blocked due to the collider~\cite{pearl2012calculus} $X$ (\ie, with two incoming arrows).
3) This makes $R\to X \to Y$ the only unblocked path from $R$ to $Y$. Hence when learning to predict $Y$ from $R$, our detector captures the causal effect free from the confounding one.
The formal definition and proof are in Appendix.

\begin{table*}[t!]
\begin{subtable}[t]{0.48\linewidth}
\scalebox{0.93}{
\def\arraystretch{1.1}
\begin{tabular}{p{2.5cm}p{1.2cm}<{\centering}p{1cm}<{\centering}p{1.2cm}<{\centering}p{1.2cm}<{\centering}}
\hline
\hline
Proposals   & K-mAP & U-R  & WI & A-OSE    \\ \hline
Selective search &57.3&  9.1& 0.029& 5,433\\
RPN & 59.0 & 8.6 &0.031  & 5,128 \\
Query based &59.7&  9.1& 0.043& 4,679\\
\textbf{Random (Ours)}  &   \cellcolor{mygray}\textbf{61.8}     &\cellcolor{mygray}\textbf{10.6}   & \cellcolor{mygray}\textbf{0.024}    & \cellcolor{mygray}\textbf{4,498}   \\

\hline
\hline
\end{tabular}
}

\parbox{8cm}{
    \caption{\label{tab:2a}\textbf{Ablation on proposal generation method}.}
}
\vspace*{-5mm}
\end{subtable} 
    \begin{subtable}[t]{0.48\linewidth}
\scalebox{0.85}{
\def\arraystretch{0.88}
\begin{tabular}{p{1.5cm}p{1cm}p{1.2cm}<{\centering}p{1cm}<{\centering}p{1.2cm}<{\centering}p{1.2cm}<{\centering}}
\hline
\hline
distribution    &num      & K-mAP & U-R  & WI & A-OSE    \\ \hline
Uniform & 100 & 60.2    &  8.7  & 0.027 & 4,877  \\
Gaussian  & 100 &   60.7     &9.3   & 0.026    & 4,732   \\
Uniform  & 500 &     \textbf{62.0}    &    10.1   & 0.026 & 4,588 \\
\cellcolor{mygray}Gaussian  & \cellcolor{mygray}500 &    \cellcolor{mygray}61.8  & \cellcolor{mygray}\textbf{10.6} & \cellcolor{mygray}\textbf{0.024} & \cellcolor{mygray}\textbf{4,498} \\
Uniform  & 1,000 &        61.7    &    10.4 & 0.028 & 4,975    \\
Gaussian  & 1,000 &        61.8    &    10.6  & 0.030 & 4,763   \\
\hline
\hline
\end{tabular}
}
\parbox{8cm}{
        \caption{\textbf{Strategy for Random Proposals}. 
        \label{tab:2b}}}
    \end{subtable} 
\hspace{20.1in}

    \begin{subtable}[t]{0.48\linewidth}
    \centering
\scalebox{0.93}{
\def\arraystretch{0.9}
\begin{tabular}{p{2.3cm}p{0.6cm}<{\centering}p{1.1cm}<{\centering}p{0.8cm}<{\centering}p{0.8cm}<{\centering}p{1.1cm}<{\centering}}
\hline
\hline
 Match &Rand   &  K-mAP & U-R  & WI & A-OSE       \\ \hline
 &   &\ 51.8 & - &0.042&7,652    \\
    Ours & & 59.7    &   9.1   &0.043&4,679   \\
    Mean activation& \Checkmark&   60.3 & 7.1 &0.024&5,929    \\
\cellcolor{mygray}Ours   &  \cellcolor{mygray}\Checkmark &\cellcolor{mygray}\textbf{61.8}   &  \cellcolor{mygray}\textbf{10.6} &\cellcolor{mygray}\textbf{0.024}&\cellcolor{mygray}\textbf{4,498}    \\
\hline
\hline

\end{tabular}
}
\parbox{8cm}{
        \caption{\textbf{Ablation on components.}\label{tab:2c}}}
    \end{subtable}
\begin{subtable}[t]{0.48\linewidth}
    \centering
\scalebox{0.95}{
\def\arraystretch{0.97}
\begin{tabular}{p{0.6cm}p{0.7cm}p{0.9cm}p{1.1cm}<{\centering}p{0.7cm}<{\centering}p{0.6cm}<{\centering}p{1.1cm}<{\centering}}
\hline
\hline
ratio&scale&location& K-mAP & U-R  & WI & A-OSE      \\ \hline
    5 &5 &20                    &     60.2      &     8.2   & 0.041 & \textbf{3,001}  \\
    10 & 10 & 20                     &   61.6  & 9.6     & 0.032 & 3,715   \\
    10 &10 &40                    & 61.3 & 10.2  & 0.029 & 4,098   \\
\cellcolor{mygray}10&\cellcolor{mygray}10&\cellcolor{mygray}100 &\cellcolor{mygray}\textbf{61.8}  & \cellcolor{mygray}\textbf{10.6}   & \cellcolor{mygray}\textbf{0.024} & \cellcolor{mygray}4,498  \\
\hline
\hline
\end{tabular}
}
\parbox{8cm}{
        \caption{\textbf{Ablation on inference bboxes.}\label{tab:2d}}}
        \label{augmentation}
    \end{subtable}
\hspace{20.1in}

    \caption{\textbf{Ablation results}. Models are trained on Task 1. We adopted the best performing setting as our default (in gray).}
    \label{ab}

\end{table*}
\section{Experiments}
\label{sec:5}

We introduce important experiment setups below, and leave other details in Appendix.

\noindent\textbf{Datasets}. We extensively evaluated RandBox on three tasks: {OWOD}, {transfer learning} and {open-set detection}.
For \textbf{OWOD}, we used Pascal VOC~\cite{everingham2010pascal} and MS-COCO~\cite{lin2014microsoft} following the standard setups~\cite{joseph2021towards,gupta2022ow}. In Task 1, the known set $K$ contains the 20 classes in Pascal VOC, and the unknown set $U$ contains the 60 classes unique to MS-COCO. For each of the Task 2-4, 20 classes in $U$ are added to $K$, \ie, Task 4 corresponds to closed-set detection with 80 classes. The training and testing images are selected from the respective split of the two datasets.
For \textbf{transfer learning}, we evaluated models trained on Task 1 on the 4,810 validation images of LVIS v1.0~\cite{gupta2019lvis}, which are from over 1000 classes that exhibits long-tailed distribution.
For \textbf{open-set detection}, we followed~\cite{miller2018dropout} to train models on Pascal VOC train split, and evaluate them on WR1~\cite{miller2018dropout}.

\noindent\textbf{Evaluation Metrics:}
On known classes, we used the standard Known-class mAP (K-mAP). On unknown classes, we used three metrics: 1) Unknown class Recall (U-R) is the recall of unknown classes at IoU threshold of 0.5. 2) Wilderness Impact (WI)~\cite{dhamija2020overlooked} is computed by $(\textrm{closed-set precision}) / (\textrm{open-set precision}) - 1$. A small WI is desired in OWOD, \ie, precision does not drop when unknown objects are introduced in evaluation. 3) Absolute Open-Set Error (A-OSE)~\cite{miller2018dropout} 
is the predictions that are mis-classified as one of the known classes.

\noindent\textbf{Implementation Details:}
In training, we used the ResNet-50~\cite{he2016deep} pre-trained on ImageNet~\cite{deng2009imagenet} as our backbone. We trained RandBox using AdamW~\cite{loshchilov2017decoupled} optimizer with learning rate as $2.5e^{-5}$ and weight decay as $1e^{-4}$.  We used the same methodology as prior work~\cite{joseph2021towards,amemiya1986instrumental,wang2020frustratingly,knoblauch2020optimal} to mitigate forgetting, \ie, by storing a balanced set of exemplars for model fine-tuning after each incremental step.
In inference, after getting the predictions on the 10,000 pre-defined bboxes, we used Non-Maximum Suppression (NMS)~\cite{neubeck2006efficient} to prune redundant bboxes. After pruning, the final predictions are the bboxes whose probability of belonging to a class in $K\cup \{``unknown''\}$ are at least 0.1.

\subsection{Main Results}
\label{sec:5.1}
Table~\ref{tab:1} presents the results of OWOD evaluation. We have the following observations:

\noindent\textbf{1)} For the conventional framework Faster-RCNN~\cite{ren2015faster} and Deformable-DETR (DDETR)~\cite{zhu2020deformable}, their K-mAP drops significantly in Task 2-4, showing that they suffer from catastrophic forgetting when new classes are incrementally added. By using data replay~\cite{joseph2021towards} (+Finetuning), their K-mAP results improve significantly (\eg, 6.0 to 29.4 on Task 4 for DDETR). Note that they are not capable to detect unknown objects (\eg, U-R is not applicable).

\noindent\textbf{2)} By comparing existing OWOD methods with the conventional frameworks, we notice that \emph{all} of them have reduced K-mAP (\eg, the DDETR-based OW-DETR reduced 1.6 mAP on Task 4 compared with DDETR+Finetuning). This means that their capabilities of unknown detection are at the cost of known detection accuracy.

\noindent\textbf{3)} By comparing RandBox with existing OWOD methods on U-R and A-OSE, we observe significant improvements, \eg,  3.1\% U-R impreovement and 391 A-OSE reduction on Task 1. This suggests that our training scheme with random boxes and the proposed matching score encourage proposal explorations, which detects and learn on more unknown objects to improve their evaluation performance.

\noindent\textbf{4)} In particular, by comparing K-mAP and WI of RandBox and OWOD methods, we see that RandBox also achieves the state-of-the-art (SOTA) performance, \eg, absolute gains ranging from 2.4\% to 8.6\% in terms of K-mAP and relatively reduces ranging from 42\% to 64\% in terms of WI. Remarkably, our RandBox even outperforms the conventional detection frameworks, \eg, Faster-RCNN and DDETR. This means that instead of sacrificing ``known'' to detect ``unknown'', RandBox learns ``unknown'' to benefit ``known''. This strongly validates that our approach removes confounding bias by simulating Randomized Controlled Experiment with random boxes, where our detector captures the causal effect from \emph{any} region proposal (known, unknown or BG) to its label (Section~\ref{sec:4}). Hence we observe unconditional improvements overall. We can also achieve SOTA on previously/currently known scores, more details are in Appendix.

\begin{figure}[t!]
\resizebox{0.98\linewidth}{40mm}{
\centering

\includegraphics[]{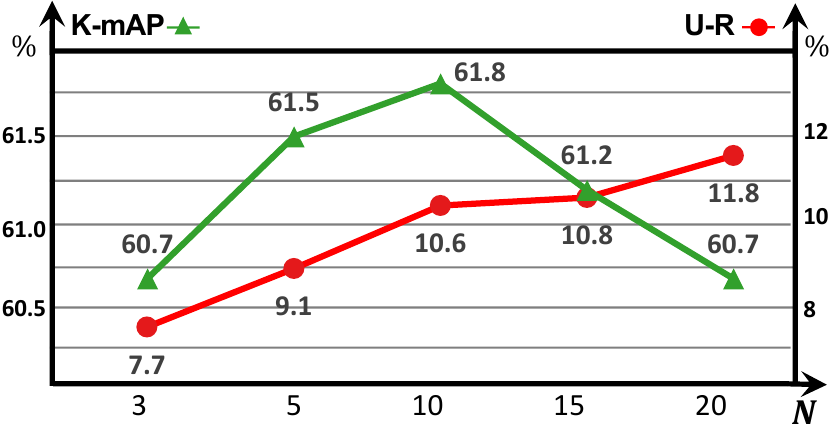}
\centering
}
\caption{Ablation on $N$ by training RandBox on Task 1.}
\label{fig:6}
\end{figure}

\subsection{Ablation Experiments}
\label{sec:5.2}

\noindent\textbf{Choice of $N$}. Recall that in training, we identify the top $N$ proposals with the largest matching score to form Unknown-FG (by removing those already in Known-FG). Hence $N$ is directly correlated with the size of Unknown-FG. The ablation results on $N$ are given in Figure~\ref{fig:6}.
Initially, increasing $N$ benefits both known detection (K-mAP) and unknown recall (U-R). This is because 1) With more unknown proposals for optimizing Eq.~\eqref{eq:3}, the detector improves in recognizing ``unknown'' (increased U-R).
2) To differentiate ``unknown'' from ``known'', the detector also avoids spurious correlations of known classes in training (\ie, captures causal effect), which may not generalize to known objects in testing (increased K-mAP).
However, further increasing $N$ leads to the drop in K-mAP. This is because increasing the size of Unknown-FG by large $N$ essentially increases the weight of the classification loss on ``unknown'' in Eq.~\eqref{eq:2}. Hence the detector learning on known classes is adversely impacted when the weight is too large.
We used $N=10$ in all our experiments.

\noindent\textbf{Proposal generation methods}
Its ablation is in Table~\ref{tab:2a}. We use our matching score Eq.~\ref{eq:4} and fix the same proposal number (500) to ablate the proposal generation methods. We compared our Random Proposals with Selective search, RPN and Query based proposals. We can find that our Random proposals surpass all other methods which means the randomness can make more explorations on more image regions, without biased towards the training data.

\noindent\textbf{Strategy for Random Proposals}. When generating random proposals in training, we experimented with different distribution (for drawing random numbers) and the number of random proposals (num), and the results are in Table~\ref{tab:2b}.
Using the same number of proposals, the results on the two distributions are both competitive.
Regarding the number of proposals, we observe that increasing it beyond 500 does not bring any improvement. This shows that the image locations are already sufficiently explored using 500 random bboxes.
Overall, we used 500 random proposals drawn from Gaussian distribution.

\noindent\textbf{Components}.
The main components in RandBox are the use of random proposals (Rand), and the matching process for Known-FG and Unknown-FG (Match). Their ablations are given in Table~\ref{tab:2c}.
The first line corresponds to the baseline Sparse R-CNN~\cite{sun2021sparse}.
Comparing the second line, by combining our matching process with NMS, our matching score can improve all metrics and already achieved SOTA (\eg, by comparing with other OWOD methods in Table~\ref{tab:1} Task 1). In the third and forth lines, we introduce the Random proposals and compare two kinds of matching score, \ie, Mean activation and our matching score. We can find that our matching score improve all metrics significantly (\eg, 1.5\% on K-mAP and 3.5\% on U-R) compared with the Mean activation. This strongly validates the effectiveness of each component in RandBox.

\begin{figure}[t!]
\resizebox{0.98\linewidth}{34mm}{
\centering

\includegraphics[]{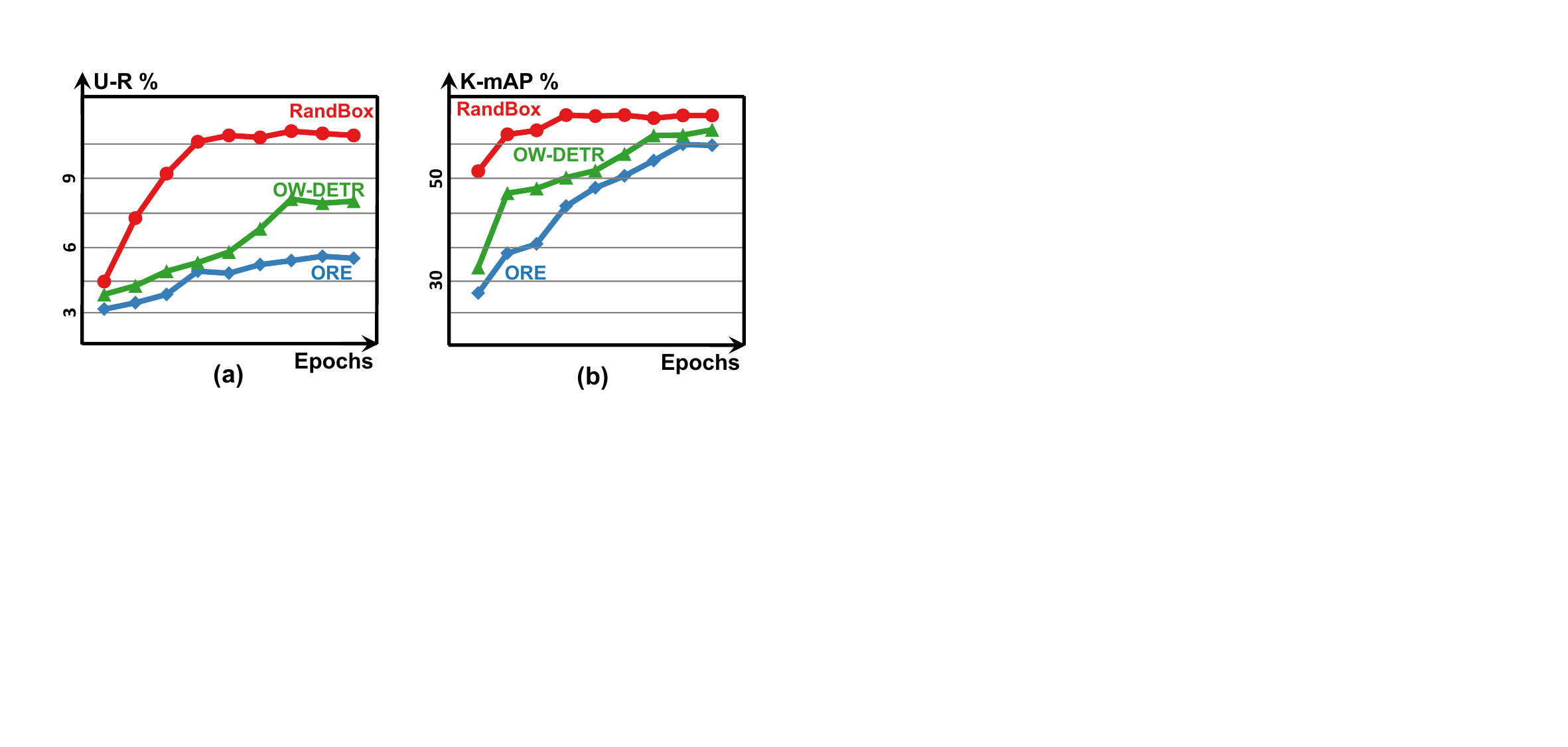}
\centering
}
\caption{Training schedule compared with existing works.}
\label{fig:7}
\end{figure}

\noindent\textbf{Training schedule}.
We show the change of U-R and K-mAP over the training and compare RandBox with ORE and OW-DETR in Figure~\ref{fig:7}. Our RandBox converges quickly and trains stably, outperforming the two methods.

\noindent\textbf{Inference Bboxes}. We ablation on the number of pre-defined bbox ratios, scales and locations in Table~\ref{tab:2d}. Increasing the bbox number generally improves the performance due to the increased recall on both known and unknown classes. However, it also increases the number of mis-classified unknown predictions, leading to a larger A-OSE.
As the performance tends to saturate with 10,000 bboxes (last row), we use this configuration for efficiency.

\subsection{Transfer and Open-Set Evaluations}
\label{sec:5.3}

\noindent\textbf{Transfer to LVIS v1.0}. We evaluated the models trained on Task 1 on LVIS 1.0.~\cite{gupta2019lvis} We observe significant improvements on all metrics compared to ORE and OW-DETR in Table~\ref{tab:3} This validates the robustness of RandBox over the existing works under distribution shift.

\begin{table}[t!]
\centering
\scalebox{0.9}{
\def\arraystretch{0.9}
\begin{tabular}{p{2.3cm}p{1.1cm}<{\centering}p{1.1cm}<{\centering}p{1.1cm}<{\centering}p{1.1cm}<{\centering}}
\hline
\hline
      Method           & K-mAP & U-R & WI   & A-OSE \\ \hline
ORE~\cite{joseph2021towards}              &   23.2   &   3.5      &  0.1743         &    23,225       \\ 
OW-DETR~\cite{gupta2022ow}                &   25.1  &    6.3       &  0.1548         &    19,672       \\ 
\textbf{RandBox} & \cellcolor{mygray}\textbf{27.7}    & \cellcolor{mygray}\textbf{8.4} & \cellcolor{mygray}\textbf{0.0735} & \cellcolor{mygray}\textbf{7,842} \\ \hline
\hline
\end{tabular}
}
\caption{Transfer evaluation results on LVIS v1.0~\cite{gupta2019lvis}.}
\label{tab:3}
\end{table}

\begin{table}[t!]
\centering
\scalebox{0.83}{
\def\arraystretch{0.88}
\begin{tabular}{p{3.3cm}p{2.5cm}<{\centering}p{2.4cm}<{\centering}}
\hline
\hline
Evaluated on →        & Pascal-VOC & Open-Set (WR1) \\ \hline
Faster R-CNN~\cite{ren2015faster} & 81.8            & 77.1           \\
RetinaNet~\cite{lin2017focal}    & 79.2            & 73.8           \\
Dropout Sampling~\cite{miller2018dropout}      & 78.1            & 73.8           \\
ORE~\cite{joseph2021towards}                   & 81.3            & 71.7           \\
OW-DETR~\cite{gupta2022ow}               & 82.1            & 78.6           \\
\textbf{RandBox}      & \cellcolor{mygray}\textbf{84.4}   & \cellcolor{mygray}\textbf{82.0}  \\ \hline \hline
\end{tabular}
}
\caption{Results on Pascal-VOC~\cite{everingham2010pascal} (closed-set) and WR1~\cite{miller2018dropout} (open-set).}
\label{tab:4}
\end{table}

\noindent\textbf{Open-Set Object Detection}. Detectors in OWOD can naturally tackle open-set detection. We followed~\cite{miller2018dropout} and evaluated models trained on Task 1 on Pascal-VOC test split (closed-set) and WR1 (open-set). We significantly improve existing SOTAs on both evaluations (Table~\ref{tab:4}).

\subsection{Additional Qualitative Results}
\label{sec:5.4}
\begin{figure}[t!]
\resizebox{\linewidth}{!}{
\centering

\includegraphics[]{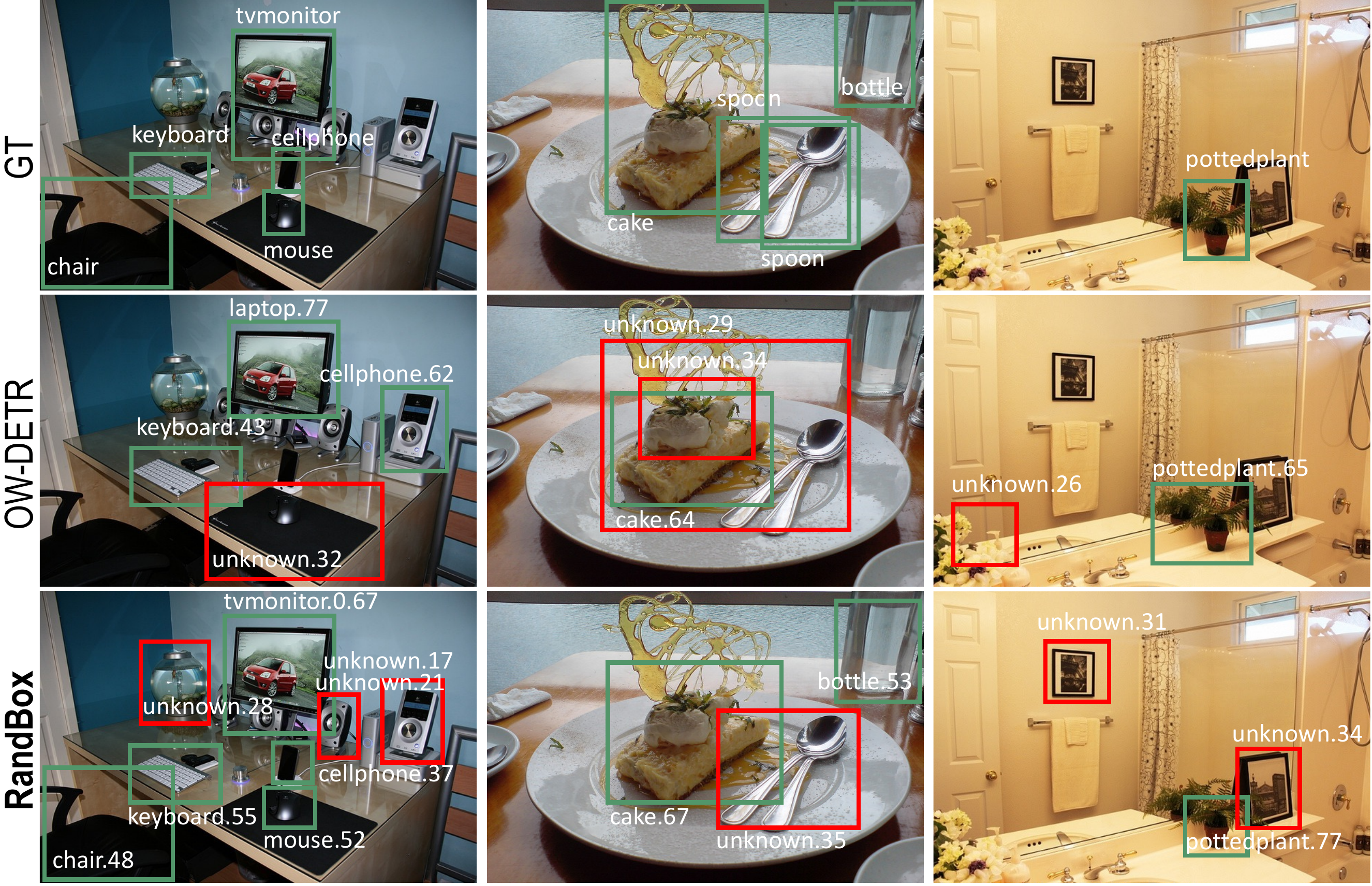}
\centering
}
\caption{Additional qualitative results. We show the ground-truth labels of known objects in the first row, and compare the predictions from OW-DETR and our RandBox. \textcolor{green}{Green}: known. \textcolor{red}{Red}: unknown.}
\label{fig:8}
\end{figure}

As shown in Figure~\ref{fig:8} first column, OW-DETR misses the known objects ``mouse'' and ``chair'' due to confusing background and partial view, respectively. Our RandBox not only detects all known objects, but also identify many valid unknown objects.
In the second column, RandBox predicts the overlapping spoons as ``unknown'', which is indeed plausible without other prior knowledge.
The third column shows that our RandBox detects the unknown object ``picture frame" both in the real world and in the mirror.

\section{Conclusion}

We presented a novel Open-World Object Detection (OWOD) method called RandBox, trained on random proposals and surpassing existing Faster R-CNN and Transformer based OWOD.
We show from a causality theoretical perspective that RandBox simulates a Randomized Controlled Experiment, which removes the confounding effect of the training data with limited known object labels.
In addition, we propose a matching score without penalizing random proposals that are not matched with the known objects, which further encourages proposal explorations.
Extensive evaluations of our proposed RandBox show that it significantly improves both the accuracy of the known objects and the recall of unknown ones.
As future direction, we will seek additional inductive bias to remove the confounding effect and extend our approach to other detection tasks.
\section{Acknowledgements}
The authors would like to thank all reviewers and ACs for their constructive suggestions. This research is supported by the National Research Foundation, Singapore under its AI Singapore Programme (AISG Award No: AISG2-RP-2021-022) and Alibaba-NTU Singapore Joint Research Institute.


\renewcommand{\thesection}{A.\arabic{section}}
\renewcommand*{\thesubsection}{A.\arabic{section}.\arabic{subsection}}
\renewcommand{\theequation}{A\arabic{equation}}
\renewcommand{\thetable}{A\arabic{table}}
\renewcommand{\thefigure}{A\arabic{figure}}

\newcommand{\norm}[1]{\left\lVert#1\right\rVert}
\newcommand\independent{\protect\mathpalette{\protect\independenT}{\perp}}
\def\independenT#1#2{\mathrel{\rlap{$#1#2$}\mkern2mu{#1#2}}}








\clearpage
\renewcommand{\thesection}{A.\arabic{section}}
\renewcommand*{\thesubsection}{A.\arabic{section}.\arabic{subsection}}
\renewcommand{\thetable}{A\arabic{table}}
\renewcommand{\thefigure}{A\arabic{figure}}
\setcounter{section}{0}
\setcounter{figure}{0}
\setcounter{table}{0}

\noindent\textbf{\Large Appendix}
\vspace{0.1in}

This appendix is organized as follows:

\begin{itemize}
    \item Section~\ref{sec:a1} gives the details of existing OWOD methods for Section 3.1, where we first give the details of detectors in Section~\ref{sec:a11}, and then give the Known-FG matching method in Section~\ref{sec:a12}. Finally, we give the details of $\mathcal{L}_{cls}$ and $\mathcal{L}_{reg}$ for Eq.~\ref{eq:1}, in Section~\ref{sec:a13}.
    \item Section~\ref{sec:a2} gives the details of 10,000 pre-defined bboxes and how we do postprocessing in the inference stage.
    \item Section~\ref{sec:a3} gives the formal definition and proof of causalities for Section 4.
    \item Section~\ref{sec:a4} provides some experiment implementation details of datasets splitting, evaluation metrics and training schedule.
    \item Section~\ref{sec:a5} shows additional results of experiments. We first give some unknown-class instances detection results by using our matching score in Section~\ref{sec:a51}. Then we show some Open-world detection qualitative results in each task in Section~\ref{sec:a52}.
\end{itemize}
\section{Preliminaries Details}
\label{sec:a1}

In this section, we first provide the details of Faster-RCNN and DETR which are frameworks of existing OWOD methods. Then we give a more detailed discussion on the Known-FG matching. Finally, we give the details of $\mathcal{L}_{cls}$ and $\mathcal{L}_{reg}$ used in training loss.

\subsection{Detectors of Existing OWOD Methods}
\label{sec:a11}
Faster R-CNN is composed of two modules. The first module is a deep fully convolutional network that proposes regions, and the second module is the Fast R-CNN detector that uses the proposed regions. The entire system is a single, unified network for object detection. Using the recently popular terminology of neural networks with ‘attention’ mechanisms, the RPN module tells the Fast R-CNN module where to look.
For DETR, two ingredients are essential for direct set predictions in detection: (1) a set
prediction loss that forces unique matching between predicted and ground truth boxes.  (2) an architecture that predicts (in a single pass) a set of objects and models their relation.
\subsection{Known-FG Matching}
\label{sec:a12}
The Known-FG matching is how to match the proposals with the ground truth. For Faster R-CNN, proposals are considered as Known-FGs if they have IoUs exceeding a threshold with ground truth. In that case, each ground truth can match multiple proposals. 
For DETR, it denote by $\{(b_i, y_i)\}_{i=1}^N$ the ground truth set of objects ($N$ is the number of objects in the image), and $\{(\hat{b}_i, \hat{y}_i)\}_{i=1}^M$ the set of $M$ predictions. Assuming $M$ is larger than $N$, then pad ground truth as a set of size $M$ (\ie, $\{(b_i, y_i)\}_{i=1}^M$) with \o (no object). To find a bipartite matching between these two sets, search for a permutation of $M$ elements $\sigma$ with the lowest cost:
\begin{equation}
    \mathop{\mathrm{argmin}}_{\sigma}\sum_{i=1}^{M}\mathcal{L}_{match}((b_i, y_i), (\hat{b}_{\sigma(i)}, \hat{y}_{\sigma(i)}))
    \label{eq:a1}
\end{equation}

where $\mathcal{L}_{match}((b_i, y_i), (\hat{b}_{\sigma(i)}, \hat{y}_{\sigma(i)}))$ is a pair-wise matching cost between ground truth $(b_i, y_i)$ and a prediction with index $(\hat{b}_{\sigma(i)}, \hat{y}_{\sigma(i)})$.
Each element $i$ of the ground truth set can be seen as a $(b_i, y_i)$ where $y_i$
is the target class label (which may be \o) and $b_i \in {[0, 1]}^4$ is a vector that defines ground truth box center coordinates and its height and width relative to the image size. For the
prediction with index $\sigma(i)$, they define probability of class $c_i$ as $\hat{y}_{\sigma(i)}(c_i)$ and the predicted box as $\hat{b}_{\sigma(i)}$. The $\mathcal{L}_{match}$ is defined as:
\begin{equation}
\mathcal{L}_{match} =
    -\mathbbm{1}_{c_i\neq\o}\hat{y}_{\sigma(i)}(c_i) + \mathbbm{1}_{c_i\neq\o}\mathcal{L}_{reg}(b_i, \hat{b}_{\sigma(i)}) 
    \label{eq:a2}
\end{equation}
where the first term is the cross-entropy and in the second term $\mathcal{L}_{reg}$ is:
\begin{equation}
\mathcal{L}_{reg} = \lambda_{iou}\mathcal{L}_{iou}(b_i,\hat{b}_{\sigma(i)}) + \lambda_{L_1}||b_i - \hat{b}_{\sigma(i)} ||
    \label{eq:a3}
\end{equation}
where $\lambda_{iou}$, $\lambda_{L_1} \in \mathbb{R}$ are hyperparameters. 
\subsection{Training Loss}
\label{sec:a13}

For the $\mathcal{L}_{cls}$ in the Eq. (1), Faster-RCNN uses the cross-entropy loss. 
\begin{equation}
\mathcal{L}_{cls} = -\sum{y_i\log{\hat{y_i}}}
    \label{eq:a4}
\end{equation}
where the $y_i$ is the $i_{th}$ sample's one-hot encoding of ground truth category and $\hat{y_i}$ is the probability prediction on each categories. In order to solve the problem of sample imbalance, focal loss~\cite{lin2017focal}, a improved cross-entropy, is a common loss calculation method.
\begin{equation}
\mathcal{L}_{cls} = -\sum{y_i(1-\hat{y_i})^\gamma\log{\hat{y_i}}}
    \label{eq:a5}
\end{equation}
where $\gamma$ is the focusing parameter (a non-negative number), and $(1-\hat{y_i})^\gamma$ is a modulating factor which can make the model focus more on difficult-to-classify samples during training by reducing the weight of easy-to-classify samples.

For $\mathcal{L}_{reg}$, Faster-RCNN uses the smooth L1 loss:
\begin{equation}
\mathcal{L}_{reg} = 
\left\{
\begin{array}{rcl}
|x|-0.5      &      & |x|>1\\
0.5x^2     &      & |x|<1\\
\end{array} \right.
    \label{eq:a6}
\end{equation}
where $x$ is the mean absolute error. The $\mathcal{L}_{cls}$ and $\mathcal{L}_{reg}$ of DETR are detailed in ~\ref{sec:a12}, which is the same as Eq.~\eqref{eq:a2} and Eq.~\eqref{eq:a3}

\section{Inference}
\label{sec:a2}
In this section, we will give the details of our inference stage including pre-defined bboxes and postprocessing.

\begin{figure}[t!]

\resizebox{\linewidth}{!}{
\centering

\includegraphics[]{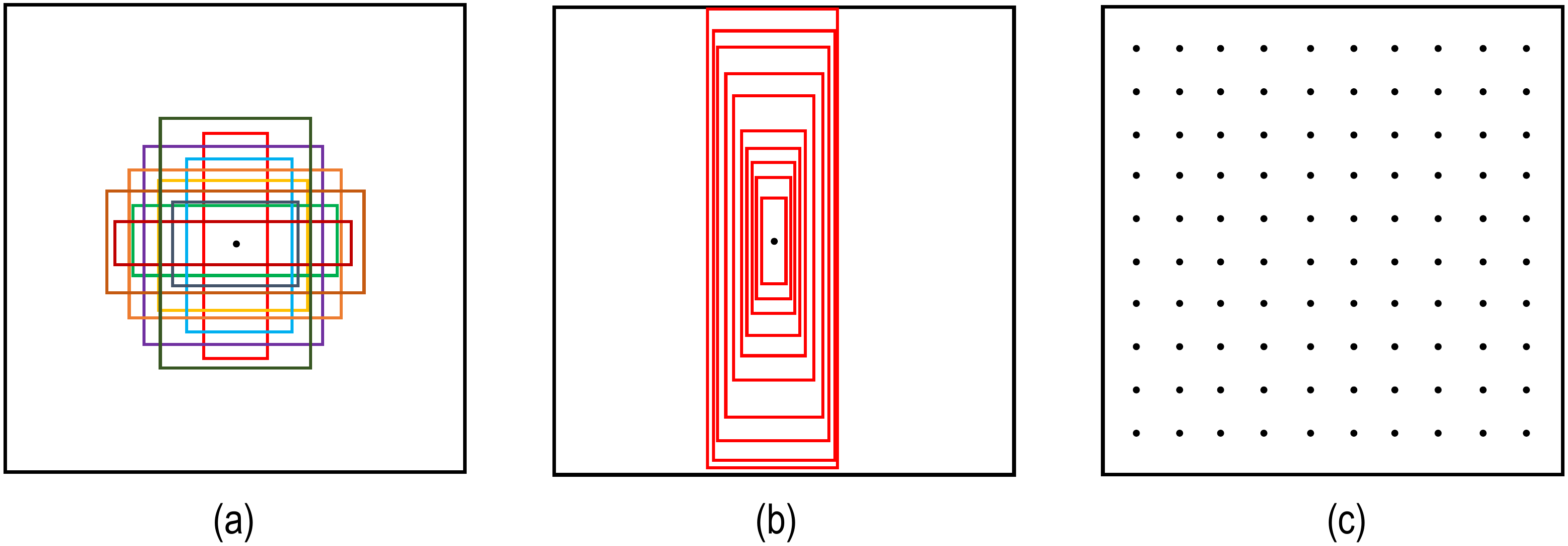}
\centering
}

\footnotesize

\caption{Our pre-defined inference bboxes.}
\label{fig:a2}
\end{figure}

\noindent\textbf{Pre-defined bboxes}. For each image, we have 10,000 pre-defined bboxes as input for inference. The 10,000 pre-defined bboxes are composed with 100 spatial locations, 10 scales and 10 aspect ratios. Specifically, for each bbox, we first evently drawn from $[0,1]$ (stride is 0.1). These 4 numbers are the (center\_x, center\_y, height, width). We show them in Figure~\ref{fig:a2}. The Figure~\ref{fig:a2}a shows that we pre-defined 10 aspect ratios when fixing the position and scale. In the Figure~\ref{fig:a2}b, we fix the position and aspect ratio, and pre-defined 10 scales. In the Figure~\ref{fig:a2}c, we give the 100 pre-defined enter coordinates of the bboxes.

\noindent\textbf{Postprocessing}. As we can only input 500 bboxes each time (our model is trained for 500 bboxes input), we infer 20 times on each image and merge all the results to get the predictions of the 10,000 pre-defined bboxes. Finally, we use non-maximum suppression to prune the bboxes at IoU threshold of 0.6.

\section{Theory}
\label{sec:a3}

In this section, we will formalize the intuition of ``blocked'' in Section 4 of the main paper, and prove that RandBox learns the causal effect from $D$ to $Y$ using \textit{do}-calculus~\cite{pearl2000models}. We will first provide the definition of d-separation and instrument variable below.

\noindent\textbf{d-separation}. A set of nodes $Z$ blocks a path $p$ if and only if 1) $p$ contains a \emph{chain} $A \rightarrow B \rightarrow C$ or a \emph{fork} $A \leftarrow B \rightarrow C$ and the middle node $B$ is in $Z$; 2) $p$ contains a \emph{collider} $A \rightarrow B \leftarrow C$ such that the middle node $B$ and its descendants are not in $Z$. If conditioning on $Z$ blocks every path between $X$ and $Y$, we say $X$ and $Y$ are \emph{d-separated} conditional on $Z$, \ie, $X$ and $Y$ are independent given $Z$ ($X \independent Y | Z$).

\noindent\textbf{Instrumental Variable}. For a structual causal model $\mathcal{G}$, a variable Z is an \emph{instrumental variable} (IV) to $X \rightarrow Y$ by satisfying the graphical criteria~\cite{pearl2009causality}: 1) $(Z \independent Y)_{\mathcal{G}_{\overline{X}}}~$; 2) $(Z \not\independent X)_{\mathcal{G}}~$, where $\mathcal{G}_{\overline{X}}$ is the manipulated graph where all incoming arrows to node $X$ are deleted.

One can verify that $R$ is not an instrumental variable in the causal graph of existing methods (Figure~\ref{fig:5}a), as it violates the first criteria with the unlocked path $R\leftarrow D \to Y$ in $\mathcal{G}_{\overline{X}}$. In contrast, $R$ is an instrument variable in RandBox (Figure ~\ref{fig:5}b).

In \textit{do}-calculus, the average causal effect from $R$ to $Y$ is given by $P(Y|do(R))$. We will show the derivation of the backdoor adjustment for $P(Y|do(R))$ in Figure 5b using the three rules of \emph{do}-calculus~\cite{pearl1988probabilistic}, and prove that $P(Y|do(R))=P(Y|R)$, \ie, by learning to predict $Y$ from random $R$, our detector learns the causal effect.

For a causal directed acyclic graph $\mathcal{G}$, let $X, Y, Z$ and $W$ be arbitrary disjoint sets of nodes. We use $\mathcal{G}_{\overline X}$ to denote the manipulated graph where all incoming arrows to node $X$ are deleted. Similarly $\mathcal{G}_{\underline X}$ represents the graph where outgoing arrows from node $X$ are deleted. We use lower case $x,y,z$ and $w$ for specific values taken by each set of nodes: $X=x, Y=y, Z=z$ and $W=w$. For any interventional distribution compatible with $\mathcal{G}$, we have the following three rules:

\noindent\textbf{Rule 1} Insertion/deletion of observations:
\begin{equation}
\begin{split}
    &P(y|do(x),z,w)=P(y|do(x),w),\\
    &\mathrm{if} (Y \independent Z | X,W)_{\mathcal{G}_{\overline X}}
\end{split}
\end{equation}

\noindent\textbf{Rule 2} Action/observation exchange:
\begin{equation}
\begin{split}
    &P(y|do(x),do(z),w)=P(y|do(x),z, w),\\
    &\mathrm{if} (Y \independent Z | X,W)_{\mathcal{G}_{\overline X \underline Z}}
\end{split}
\end{equation}

\noindent\textbf{Rule 3} Insertion/deletion of actions:
\begin{equation}
\begin{split}
    &P(y|do(x),do(z),w)=P(y|do(x),w),\\
    &\mathrm{if} (Y \independent Z | X,W)_{\mathcal{G}_{\overline X \overline {Z(W)}}},
\end{split}
\end{equation}
where $Z(W)$ is the set of nodes in $Z$ that are not ancestors of any $W$-node in $\mathcal{G}_{\overline{X}}$.

In our causal graph (Figure 5b), the desired interventional distribution $P(Y|do(R=\mathbf{r}))$ can be derived by:
\begin{align}
    P(Y|do(\mathbf{r})) &= \sum_{d} P(Y|do(\mathbf{r}),d) P(D=d|do(\mathbf{x})) \label{bd1}\\
                        &= \sum_{d} P(Y|do(R=\mathbf{r}),d) P(D=d) \label{bd2}\\
                        &= \sum_{d} P(Y|R=\mathbf{r},d) P(D=d) \label{bd3}\\
                        &= \sum_{d} P(Y|R=\mathbf{r},d) P(D=d|R=r) \label{bd4}\\
                        &= P(Y|R=r) \label{bd5}
\end{align}
where Eq.~\eqref{bd1} and Eq.~\eqref{bd5} follow the law of total probability; Eq.~\eqref{bd2} uses Rule 3 given $D \independent X$ in $\mathcal{G}_{\overline{X}}$; Eq.~\eqref{bd3} uses Rule 2 to change the intervention term to observation as $(Y \independent X | D)$ in $\mathcal{G}_{\underline{X}}$. Eq.~\eqref{bd4} is because $D$ and $R$ are d-separated by $\emptyset$, \ie, $D\independent R$.

\section{Implementation Details}
\label{sec:a4}

In this section, we give the details of how the datasets splitted in 4 tasks, specific IoU threshold of evaluation metrics and the detailed training schedule for each task.

\noindent\textbf{Datasets Split.} The 80 classes of MS-COCO\cite{lin2014microsoft} are split into 4 tasks and the number of images as well as instances in each task are shown in Table.\ref{tab:appendix1}. 
\begin{table}[t!]
\begin{center}
\resizebox{\linewidth}{!}{
\begin{tabular}{lcccc}
\hline
\hline
                   & Task 1  & Task 2           & Task 3 & Task 4             \\ \hline
\multirow{2}{*}{Semantic split} & VOC~\cite{everingham2010pascal} & Outdoor, Accessories, & Sports, & Electronic, Indoor, \\
                   & Classes & Appliance, Truck & Food   & Kitchen, Furniture \\ \hline
train images    & 16551   & 45520            & 39402  & 40260              \\
test images     & 4952    & 1914             & 1642   & 1738               \\
train instances & 47223   & 113741           & 114452 & 138996             \\
test instances  & 14976   & 4966             & 4826   & 6039               \\ \hline \hline
\end{tabular}}
\end{center}
    \caption{Task composition in the OWOD evaluation protocol.}
    \label{tab:appendix1}
\end{table}

\noindent\textbf{Evaluation Metrics}. As described in Section 5, we give the specific IoU threshold for 4 metrics here. The Known-mAP (K-mAP) is at IoU threshold of 0.5, Unknown-Recall 
(U-R) is at IoU threshold of 0.5 , Wilderness Impact (WI) is at IoU threshold of 0.8 and Absolute open set error (A-OSE) is at IoU threshold of 0.5.

\noindent\textbf{Training Schedule}. For task 1, the training schedule is 20K iterations, with the learning rate divided by 10 at 15K and 18K iterations. For task 2,3,4, the training schedules are 15K iterations, with the learning rate divided by 10 at 10K iterations. After task 2,3,4, we fine-tuning the model for 15K iterations, with the learning rate divided by 10 at 5K, 10K, 12K iterations. We calculate the $\mathcal{L}^{U}$ in Eq.~\ref{eq:2} until 500 iterations in each task due to the matching score is not accurate at the beginning. All model are trained with a mini-batch size 12 on 2 A100 GPUs.

\section{Additional Results}
\label{sec:a5}
In this section, we first give some results on unknown-class detection in~\ref{sec:a51}. Then, we give the previously/currently known scores in~\ref{sec:a52}. In addition, we give some open-world detection qualitative results in~\ref{sec:a53}.

\subsection{Unknown-Class Detection}
\label{sec:a51}
In Section 5, we give 4 metrics, \ie the standard Known-class mAP (K-mAP), recall of unknown classes (U-R), Wilderness Impact (WI)~\cite{dhamija2020overlooked} and Absolute Open-Set Error (A-OSE)~\cite{miller2018dropout}. In this Section, we add a metric: standard Unknown-class AP (U-AP) at different IoU threshold (\eg, U-AP50 is the unknown-class average precision at 0.5 IoU threshol and U-AP is the average of U-AP[50,95]). This can more intuitively measure both the recall and precision of unknown-class instances.
We calculate the $\mathcal{L}^{U}$ in Eq.~\ref{eq:2} until 500 iterations in each task due to the matching score is not accurate at the beginning. We call this operation 'warm up'. In Table~\ref{tab:appendix2}, we follow the same setting in Section 5.2 and show the results on with warm up operation or not. The results show that our warm up operation can improve the U-AP because after some warm up iteration, our matching score can recall the Unknown-FG with higer precision. Then, we show the U-AP on different inference Bboxes in Table~\ref{tab:appendix3}. The results show that increasing the number of inference bboxes generally improves the U-AP due to the increased recall on unknown-class instances.
\begin{table}[t!]
\centering
\scalebox{0.95}{
\def\arraystretch{1.05}
\begin{tabular}{p{1.3cm}p{1.3cm}<{\centering}p{1.3cm}<{\centering}p{1.3cm}<{\centering}}
\hline
\hline
warm up& U-AP & U-AP50  & U-AP75       \\ \hline
                    &     1.98      &     4.46   & 1.57  \\
\cellcolor{mygray}\checkmark                    &   \cellcolor{mygray}\textbf{2.57}  & \cellcolor{mygray}\textbf{5.25} & \cellcolor{mygray}\textbf{2.23}   \\
\hline
\hline
\end{tabular}
}
\caption{Unknown-class Average precision (U-AP) on whether to use warm up operation.}
\label{tab:appendix2}
\end{table}
 
\begin{table}[t!]
\centering
\scalebox{0.95}{
\def\arraystretch{1.05}
\begin{tabular}{p{0.6cm}p{0.7cm}p{0.9cm}p{1.3cm}<{\centering}p{1.3cm}<{\centering}p{1.3cm}<{\centering}}
\hline
\hline
ratio&scale&location& U-AP & U-AP50  & U-AP75       \\ \hline
    5 &5 &20                    &     2.33      &     4.47   & 2.10  \\
    10 & 10 & 20                     &   2.56  & 5.20     & 2.22    \\
\cellcolor{mygray}10&\cellcolor{mygray}10&\cellcolor{mygray}100 &\cellcolor{mygray}\textbf{2.57}  & \cellcolor{mygray}\textbf{5.25}   & \cellcolor{mygray}\textbf{2.23}   \\
\hline
\hline
\end{tabular}
}
\caption{Unknown-class Average precision (U-AP) on different inference bboxes.}
\label{tab:appendix3}
\end{table}

\subsection{Previously/Currently Known Scores}
\label{sec:a52}
In Table~\ref{tab:appendix4}, we give the previously/currently known details as supplementary to Table 1.
\begin{table}[t]
\scalebox{0.82}{
\def\arraystretch{1.1}
\begin{tabular}{p{0.7cm}p{1.1cm}<{\centering}p{0.7cm}<{\centering}p{0.1cm}<{\centering}p{1.1cm}<{\centering}p{0.7cm}<{\centering}p{0.1cm}<{\centering}p{1.1cm}<{\centering}p{0.7cm}<{\centering}}
\hline
\hline
\multirow{2}{*}{$\beta$} & \multicolumn{2}{c}{ORE~\cite{joseph2021towards}} & &\multicolumn{2}{c}{OW-DETR~\cite{gupta2022ow}} &&\multicolumn{2}{c}{RandBox} \\ \cline{2-3} \cline{5-6} \cline{8-9}
                      & K-mAP         & U-R   & &K-mAP         & U-R      & &K-mAP         & U-R         \\ \hline
0.05          &  55.8 & 4.3  & &59.3  & 6.9  & & 61.3   &    9.4         \\
\cellcolor{mygray}0.1                   &  56.0     &   4.9 &&    59.3         &  7.5            &&   \cellcolor{mygray}\textbf{61.8}  &     \cellcolor{mygray}10.6   \\
0.2     &  56.4    &  5.3   & &58.5 & 8.2    & &      60.2         &  11.5    \\
0.5     &  56.2    &  5.8   &&57.6 &   9.1   & &      59.5         &  \textbf{11.6}    \\
 \hline
\hline
\end{tabular}
}
        \caption{Additional results supplementary to Table 1.\label{tab:appendix4}}
    \end{table}
\subsection{Open-world Detection Qualitative Results}
\label{sec:a53}
\begin{figure}[t!]

\resizebox{\linewidth}{!}{
\centering

\includegraphics[]{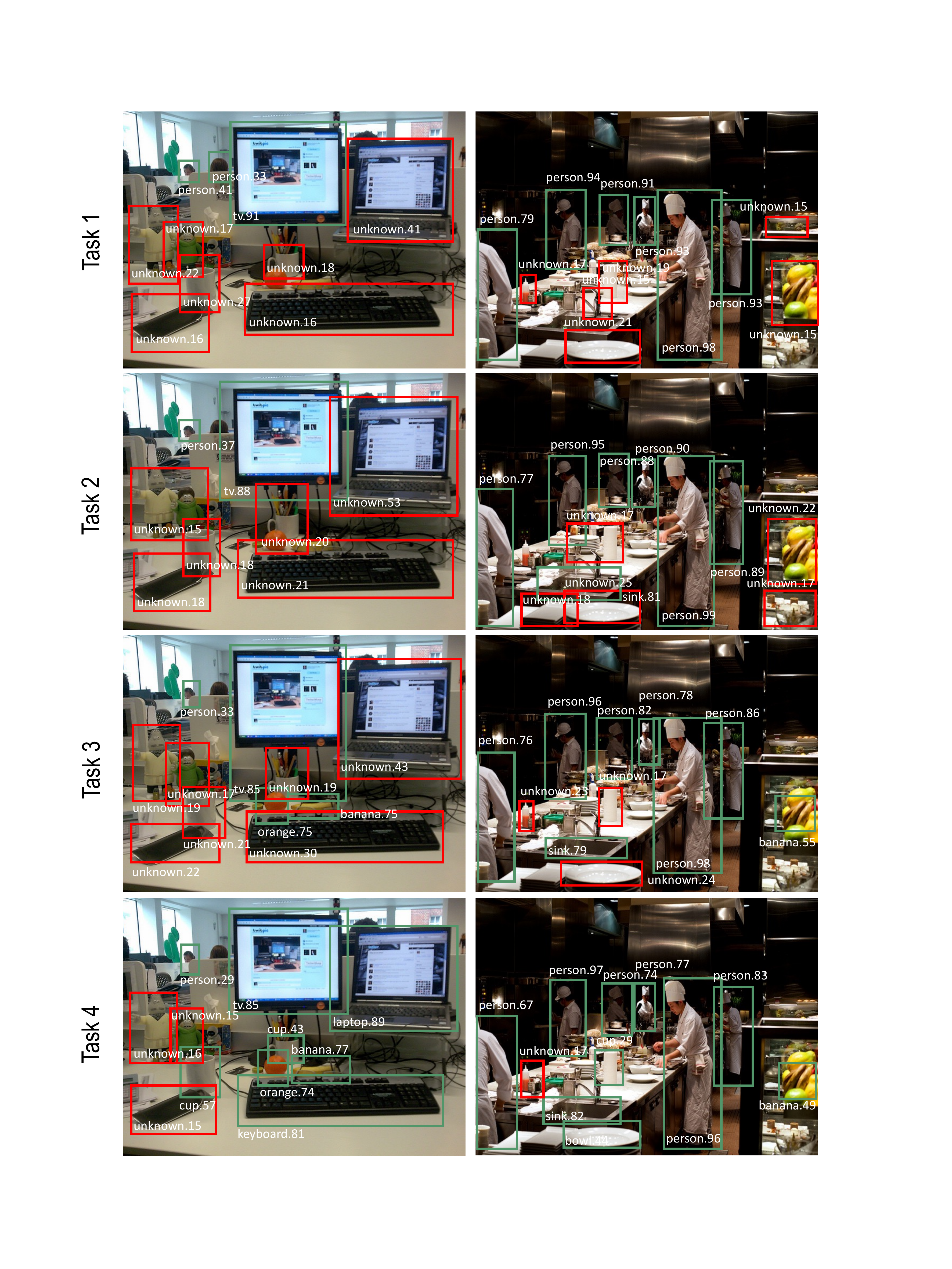}
\centering
}

\footnotesize

\caption{\textbf{Open-world Detection Qualitative Results. \textcolor{green}{Green}: known. \textcolor{red}{Red}: unknown.}}
\label{fig:a1}
\end{figure}

We give some open-world detection qualitative results in Figure~\ref{fig:a1}. This figure show a category incremental learning process from top to bottom (Task 1 to Task 4). Two columns are two cases.

First, we look at the first column (case 1).
We can see the first row (Task 1), RandBox detects 'tv', 'person' and multiple 'unknown' instances. In these 'unknown' instances, we can known they are 'keyboard', 'cup', 'laptop' and so on. These 'unknown' instances will be detected as known-class instances as the increasing task number.  And in the second row (Task 2), there are still 'tv' and 'person' are known classes. In the third row (Task 3), 'orange' and 'banana' were detected as known class after they were introduced to RandBox in Task 3. Finally, in the forth row (Task 4), 'keyboard', 'laptop' and 'cup' were detected as known class.
Similarly, we can see another case in the second column. In the first row (Task 1), there are some 'person' are detected as known-class instances and multiple 'unknown' instances. In these 'unknown' instances, we can known they are 'sink', 'banana', 'bowl' and so on. In the second row (Task 2), the 'sink' was detected as known class besides 'person'. In the third row (Task 3), the 'banana' was detected as known class after it was introduced in Task 3. Finally, in the forth row (Task 4), RandBox detected the 'bowl' and 'cup'. These results show how our RandBox implement the Open-world detection.

{\small
\bibliographystyle{ieee_fullname}
\bibliography{egbib}
}

\end{document}